\def\year{2021}\relax
\documentclass[letterpaper]{article} 
\usepackage{aaai21}  
\usepackage{times}  
\usepackage{helvet} 
\usepackage{courier}  
\usepackage[hyphens]{url}  
\usepackage{graphicx} 
\usepackage{natbib}  
\usepackage{caption} 
\usepackage{multirow}
\usepackage{amsmath}
\usepackage{booktabs}
\usepackage{float}
\restylefloat{table}

\title{Invertible Concept-based Explanations for CNN Models with Non-negative Concept Activation Vectors}
\author {
    Ruihan Zhang,
    Prashan Madumal,
    Tim Miller,
    Krista A. Ehinger,
    Benjamin I. P. Rubinstein\\
}
\affiliations {
    School of Computing and Information Systems\\
    The University of Melbourne,  Victoria, Australia \\
    \{ruihanz, pmathugama\}@student.unimelb.edu.au, \{tmiller,kehinger,benjamin.rubinstein\}@unimelb.edu.au
}

\begin{document}

\maketitle

\begin{abstract}

Convolutional neural network (CNN) models for computer vision are powerful but lack explainability in their most basic form. This deficiency remains a key challenge when applying CNNs in important domains. Recent work on explanations through feature importance of approximate linear models has moved from input-level features (pixels or segments) to features from mid-layer feature maps in the form of concept activation vectors (CAVs).  CAVs contain concept-level information and could be learned via clustering. In this work, we rethink the ACE algorithm of Ghorbani et~al., proposing an alternative invertible concept-based explanation (ICE) framework to overcome its shortcomings. Based on the requirements of fidelity (approximate models to target models) and interpretability (being meaningful to people), we design measurements and evaluate a range of matrix factorization methods with our framework. We find that \emph{non-negative concept activation vectors} (NCAVs) from non-negative matrix factorization provide superior performance in interpretability and fidelity based on computational and human subject experiments. Our framework provides both local and global concept-level explanations for pre-trained CNN models.
\end{abstract}

\section{Introduction}

Deep learners such as convolutional neural networks (CNNs)~\cite{he2016deep} are widely used across important domains like computer vision due to demonstrated performance in numerous tasks. However, when applying to critical domains like medicine, justice, and finance, explainability has become a key enabler and mitigation for applications. While commentators like \citet{rudin2019stop} argue that deep learning approaches should not be used for these risky domains, using deep learning to discover features for more `interpretable' models requires explainability to determine what features have been discovered.

Recent CNN explanation methods attempt to quantify the importance of each feature. Feature importance usually corresponds to a linear approximation of highly complex models. Linear models are simple and explainable with understandable features and weights. Methods like CAM~\cite{zhou2016learning}, LIME~\cite{ribeiro2016should} and saliency maps~\cite{bach2015pixel} use input-level features. However, these methods only point out the important areas in an image, rather than, for example, identifying key concepts. To make explanations more human understandable, more recent work uses features from within the CNN models with higher-level information~\cite{kim2018interpretability,bau2017network,zhou2018interpretable} under the assumption that these are closer to human-relevant concepts than areas of pixels. This work uses supervised learning to classify feature maps with concepts and use weights to indicate the directions of certain concepts called Concept Activation Vectors (CAVs) in the feature map space. To reduce the labeling workload, Automated Concept-based Explanations (ACE)~\cite{ghorbani2019towards} apply unsupervised learning to segments of images from certain classes and generate clusters of image segments as CAVs.

ACE is powerful for learning features (CAVs) using only images from target classes without any labels but has several drawbacks. First, learned concept weights are inconsistent for different instances. This is a common problem for all linear approximation explanation methods as they are all approximations to the target CNN models. Second, it is difficult to measure the performance of learned sets of CAVs. ACE is a one-way explainer from the target model to explanations and irrelevant segments are discarded. Transforming an explanation back to the target model's prediction is difficult with segments. However,  how much information is lost in the explanation? Information could be lost in unused segments and distances between used segments and their cluster centroids. 

\Citet{olah2018the} introduces an interpretable interface with many methods to display what's inside the CNN model for a given image to achieve interpretability. Feature maps can be shown using different grouping methods (axes). They also introduce a method using non-negative matrix factorization (NMF) on feature maps to gather interpretable information. This shares a similar idea with ACE as the clustering methods like $k$-means used in ACE could be considered as a matrix factorization (dimensionality reduction) method. In addition to reducing the dimensionality, matrix factorization also analyzes the information loss with inverse functions. The feature maps from segments used in ACE are replaced by spatial activations from the feature map. This replacement is also used in CoCoX~\cite{akula2020cocox}. 

\begin{figure*}[h]
    \centering
    \includegraphics[scale=0.12]{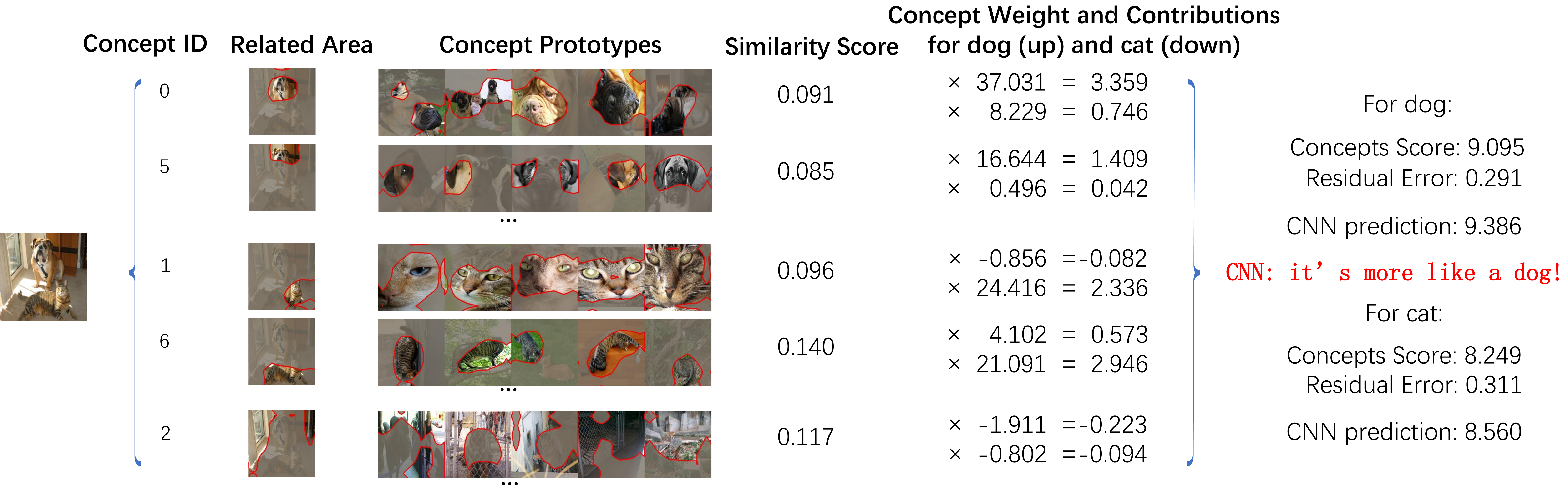}
    \caption{Explanation for an image with a dog and cat. For each concept, the framework provides prototypes based on the training set and correlates areas as global explanations. The explainer decomposes the final prediction score to concept scores and weights through a linear model to explain locally. The explanation is based on the ResNet50 CNN model from TorchVision. The full version for this explanations is shown in the appendix}
    \label{fig:exp_sample}
\end{figure*}

In this paper, we modify the ACE framework using matrix factorization for feature maps instead of clustering segments' feature maps as in ACE to overcome its shortcomings. Targeting the last feature map of CNN models, our framework could provide consistent CAVs weights for different instances (in most CNN models). Through the inverse function of matrix factorization, the information lost (fidelity, differences in prediction) in the explanations can be measured. Given performance requirements (e.g. accuracy of approximations), explainers (hyper-parameters) could then be learned automatically. Also, inverse functions allow us to analyze the distributions of contributions from learned CAVs to provide detailed local explanations for given instances. ACE only provides global explanations for classes. Having this framework, we compare three different methods $k$-means, PCA and NMF. Sample local and global explanations are shown in Figure~\ref{fig:exp_sample}. Our contributions are:

\begin{itemize}
    \item We propose a new concept-based explanation framework that provides consistent weights for features (under some limitations) and consistent fidelity measurement using non-negative matrix factorization.
    \item We propose new measurements of fidelity for CAV-based interpretability methods. This measures how accurate the explanations are to the original CNN model. Hyper-parameters can be learned through this measurement.
    \item We propose a new human subject experiment design for measuring  the interpretability of concept-based interpretability methods. Unlike previous work, we measure interpretability in a scientific manner using human studies, rather than relying on ``intuition'' and  selected examples to illustrate interpretability \cite{leavitt2020towards}.
 
\end{itemize}

\section{A Framework for Concept-based Explanations}

\begin{figure*}[ht]
    \centering
    \includegraphics[scale=0.18]{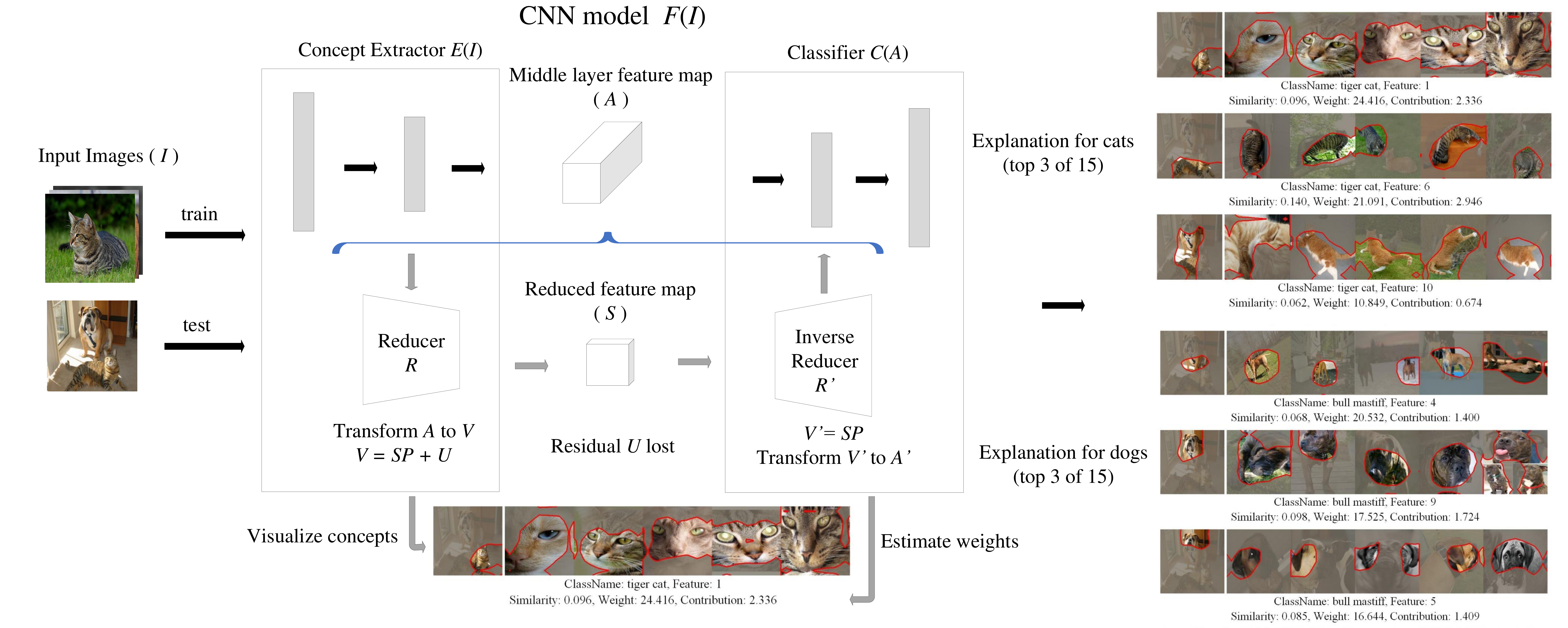}
    \caption{A depiction of our framework. The CNN model is separated into concept extractor and classifier by chosen middle layer with a reducer. The concept extractor provides concept presentations, instance correlated areas and feature scores. The classifier provides concept weights and generates linear approximations as an explanation.}
    \label{fig:framework}
\end{figure*}

Concept-based explanations may be approached as linear approximations for separate CNN models as follows. First, we separate the CNN classifier into a concept extractor and classifier from a single CNN layer, and the explanations are based on the feature maps from that layer. Input-level explanations like LIME~\cite{ribeiro2016should} and saliency maps~\cite{shrikumar2017learning,smilkov2017smoothgrad} skip this step and use input pixels as features for approximate models directly. Second, we apply matrix factorization to the feature maps to provide CAVs for the next step. A reducer is trained with a target concept-related dataset. Note that the middle-layer feature maps may contain too many dimensions, and information in each dimension is not enough to be meaningful. Therefore, the reducer may gather information separated in all dimensions to provide CAVs and reduce the complexity of the approximate model. For the final step, we build a linear approximation to the classifier and estimate the concept importance for each CAV. The explanation is based on the learned reducer and estimated weights for each CAV. For explanations of new inputs, reducers provide meaningful concept descriptions and concept scores from the feature maps. A diagram of the framework is shown in Figure~\ref{fig:framework}.

The selection of the target layer is important. Higher layers focus more on concepts (high-level features) and lower layers focus more on edges and textures of the image (low-level features)~\cite{zeiler2014visualizing}. If the reduced concepts are to be meaningful, the selection of a higher layer seems more suitable. Higher layers usually mean a classifier with fewer layers (simpler) and a concept extractor with higher-level concepts. This provides more accurate estimated weights. One special case is when using feature maps from the last layers when they are a Global Average Pooling (GAP) layer and a dense layer, the classifier under explanation will reduce to a simple linear model. Estimated weights are accurate as they are constant at any position under any CAVs. The previous layers' weight estimates take the average weights of all instances. Weights could vary for different inputs. In this paper,  we assume that the last layer is the most suitable for concept-based explanations, so we use this as the target layer.

A benefit of using reducers instead of clustering methods is that reducers provide scores for concepts as outputs instead of predictions of clusters' centroids. The reduced concept scores can be applied to the approximate model to analyze the contribution distribution for each feature more accurately. In ACE, concept scores can only be binary. Thus, reducers provide better fidelity when inverting the reduction process. This could help when evaluating the fidelity of the learned CAVs and also when applying the concepts in a larger explainability framework.

\Citet{ribeiro2016should} claim two important requirements for linear approximations: interpretability and fidelity. \textbf{Interpretability} means that the feature representation used in the approximate model needs to be meaningful to human observers. \textbf{Fidelity} prescribes that the approximate model should make similar predictions to the model under explanation. This clearly introduces a dual-objective problem: increasing fidelity can decrease interpretability, or vice-versa. In this paper, we evaluate different matrix factorization methods on these two measures.

\section{Methodology}
Given a pre-trained CNN classifier $F$ with $n$ training images $I$, the prediction process will be $F(I) = {Y}$. We  remove any final softmax layer (if present) so that each ${y}$ is a scalar instead of a probability. Let $A$ be the feature map from the target layer $l$, then $F$ can be seperated into two parts, feature extractor $E_l(I) = A_l$ and classifier $C_l(A_l) = Y$. Feature map $A$ should be of shape $n \times h \times w \times c$ where $h$ and $w$ reflect the feature map size and $c$ is the number of channels. $A$ is assumed to be non-negative as most recent CNN models use the \textit{relu} activation. Let $a^{(i,j)}$ be a vector from $A$ at position $(i,j), \{0\leq i < h,0 \leq j<w\}$. CNN models share weights, so vector $a$ at each position in $A$ could be considered as a vector description of the original images but with different correlated receptive fields after equivalent processing. 

\subsubsection{Non-Negative Concept Activation Vectors}
Feature map $A$ can be flattened to $V \in R^{\left(n \times h \times w\right) \times c}$. Non-negative matrix factorization (NMF) reduces the channel dimensions of non-negative matrix $V$ from $c$ to $c'$. Here $V$ is reduced to feature score $S \in R^{\left( n \times h \times w \right) \times c'}$ and feature direction $P \in R^{c' \times c}$ as $V = SP + U$. The aim is to minimize the residual error $U$. It is given formally as $\underset{S,P}{\min}  \|V-SP\|_F ~~~\mbox{s.t.} ~~~ S \geq 0, P \geq 0 $. Having images with the same label or concept, most vectors $a$ with different correlated receptive fields in $A$ can be considered as a vector description of a target concept part (e.g. eye, mouth or ear of a dog). Factorization on these vectors can disentangle important and frequently-appearing directions for target concepts.

$P$, the meaningful NCAVs in the feature map dimension space, is a fixed parameter for the explainer after being trained with $A$ from some images. Given the explainer, for feature maps from new images, we can apply NMF with $P$ to get $S$. $P$ is a vector basis and $S$ is the length of projections of $a$ on these directions. $S$ can be considered as feature scores for feature directions in $P$ as it's the degree to which vector $a$ is related (similar) to the NCAVs in $P$.

\subsubsection{Weight Estimation}
For weights or concept importance, other interpretability methods such as saliency maps use the gradients of output scores for some classes with respect to individual input features. Feature importance of classifier $C$ can be estimated using the method in TCAV~\cite{kim2018interpretability}.
Given a learned NCAV $p_l$ in target layer $l$,  the estimated feature importance for a classifier targeting class $k$ for given feature map $A_{l}$ is $ \frac{\partial C_{l,k}}{\partial p_l} = \frac{1}{n \times h \times w} \sum_{a \in A_l}\lim_{\epsilon\to 0} \frac{h_{l,k}(a+\epsilon p_l)- h_{l,k}(a-\epsilon p_l)}{2\epsilon}$. Classifier $C$ is usually a non-linear function. Estimated weight is an average derivative over the area sampled by $A$, and $A$ should come from images related to the target concept to reduce estimate error.

Only a linear classifier has a consistent derivative over the whole input space, so the best choice for target layer is the last layer before a GAP layer and a dense layer. Having weights $W \in R^{c}$ with bias $b$ from the last dense layer for target class $k$ and learned NCAV parameter $P$, the classifier $C$ will be:
\begin{align*}
    C_{k,l}(A) & = GAP(A)W + b\\
                   & = GAP(S)PW + GAP(U)W +b
\end{align*}
The feature importance for NCAV $P$ will be $PW$. This is independent of the input feature map $A$. 

\subsubsection{Vector Visualization}
There are many ways to visualize a vector from a layer. For instance, having a vector in a middle layer, Deep Dream concept vector visualization~\cite{olah2017feature}  provides a pattern-enhanced image based on gradients for the vector. In this work, we use the method of prototypes~\cite{kim2016examples}, choosing images containing target concepts and highlighting these concepts. Applying GAP to the decomposed feature maps, we can provide a score for each concept. Images with high concept scores are taken as the prototypes. Previous work shows that middle-layer feature maps have spatial correlations with input images, as was used in image segmentation to replace input masks with feature map masks~\cite{dai2015convolutional}. Decomposed feature maps for a single CAV could be presented as heatmaps for target concepts. Combining a heatmap and an image, we can apply a threshold for the heatmap and highlight only areas with high concept value in the image. In this paper, the threshold is taken to be 0.5, and only regions with values higher than 0.5 (after a min-max normalization) are considered to be related. Concept prototypes from Figure~\ref{fig:exp_sample} are visualized in this way.


\section{Evaluation}
Following the desiderata of this work, we aim to measure both fidelity and interpretability. Fidelity is measured computationally while for interpretability, we propose a new task that uses human subject experiments.

For both the computational and human subject experiments, we use well-known CNN models for image classification. We consider  two different datasets: ILSVRC2012 (ImageNet)~\cite{ILSVRC15} and CUB~\cite{WahCUB_200_2011}. The implementation is based on PyTorch and scikit-learn. CNN models used for the ILSVRC2012 dataset are from torchvision pre-trained models. The top1 error of ResNet50 on ILSVRC2012 is $23.85\%$ and that of Inception-V3 on ILSVRC2012 is $22.55\%$. For the CUB dataset, we use the ResNet50~\cite{he2016deep} architecture and apply fine-tuning based on ImageNet pre-trained weights. The top1 error is $15.81\%$. For comparison with NMF, we choose the baseline of clustering (from ACE) and PCA (a popular matrix factorization method). Reducers are trained based on the training set and evaluated on the test set or the validation set.

\subsection{Fidelity for Approximate Models}

In this section, we compare the fidelity of approximate models using the three different matrix factorization methods. We evaluate the fidelity for CNN pre-trained models with different $c'$ for each matrix factorization method using both classification and regression measurements. 

For fidelity, we measure the difference between the approximate and original model predictions, but the measurements for classification and regression problems are different. Classification models only focus on predicting labels, so errors that do not change the predicted labels are ignored. For regression, any difference in approximate models will affect the performance based on the loss function measurement.

\subsubsection{Measures} 
Given the original model $F$ and an approximate model $\hat{F}$, \Citet{craven1996extracting} measure  the fidelity of approximate classification models as the 0-1 loss. This targets the difference in accuracy through predictions. For regression, \citet{ribeiro2016should} measure fidelity as the squared error $(F(X) - \hat{F}(X))^2$. While the squared error is appropriate as a loss function during training, for evaluation, relative error (RE) is more easily interpretable, being scale-free. Given $F$, $\hat{F}$ and a set of images $I$, the measurement for classification and regression models based on the dataset will be:
\begin{align}
    Fid\_c_{F,\hat{F}}(I) &= \frac{\#\{i \in I \mid F(i)=\hat{F}(i)\}}{\#\{I\}} \\
    Fid\_r_{F,\hat{F}}(I) &= \frac{\sum_{i \in I}{|F(i)-\hat{F}(i)|}}{\sum_{i \in I}{|F(i)| + \epsilon}}
\end{align}
Having a trained reducer $R$ and its inverse function $R'$ for layer $l$, the approximate model is given by $\hat{F}_l(I) = C_l(R'(R(E_l(I))))$. 

\subsubsection{Experimental Setup} 
Our experiment is based on a ResNet50 pre-trained model for ImageNet from torchvision and a ResNet50 pre-trained model for CUB. Both of them use the feature maps from \textit{layer4}'s output. The parameter $c'$ is evaluated  from $5$ to $50$, in steps of $5$. The model can be considered as both a classification and regression (score for every single class) model. So both fidelity measures can be evaluated. Here we trained reducers for all 1,000 classes in ILSVRC2012 and all 200 classes in CUB. 
Reducers are trained with images from a single class. For classification methods, only the top 5 classes are considered as candidates. For regression, only ground truth classes are tested, calculating the RE for the approximate models' outputs. We take the mean RE for all classes as the final result.

\begin{figure}[tb]
    \centering
    \includegraphics[scale=0.21]{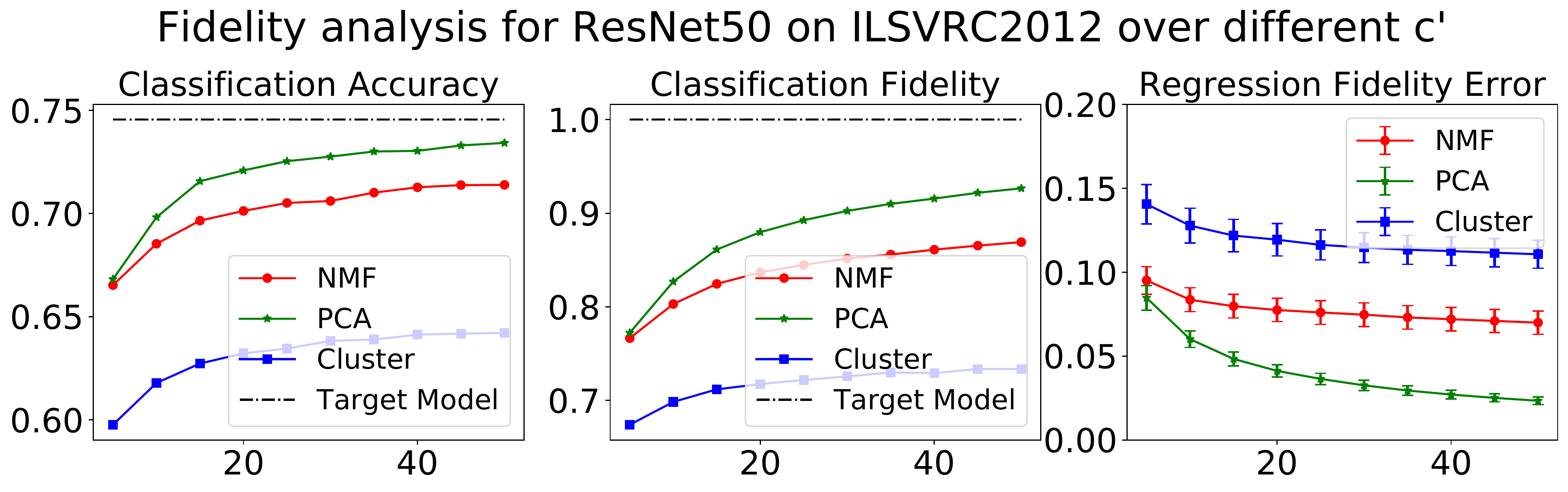}
    \includegraphics[scale=0.21]{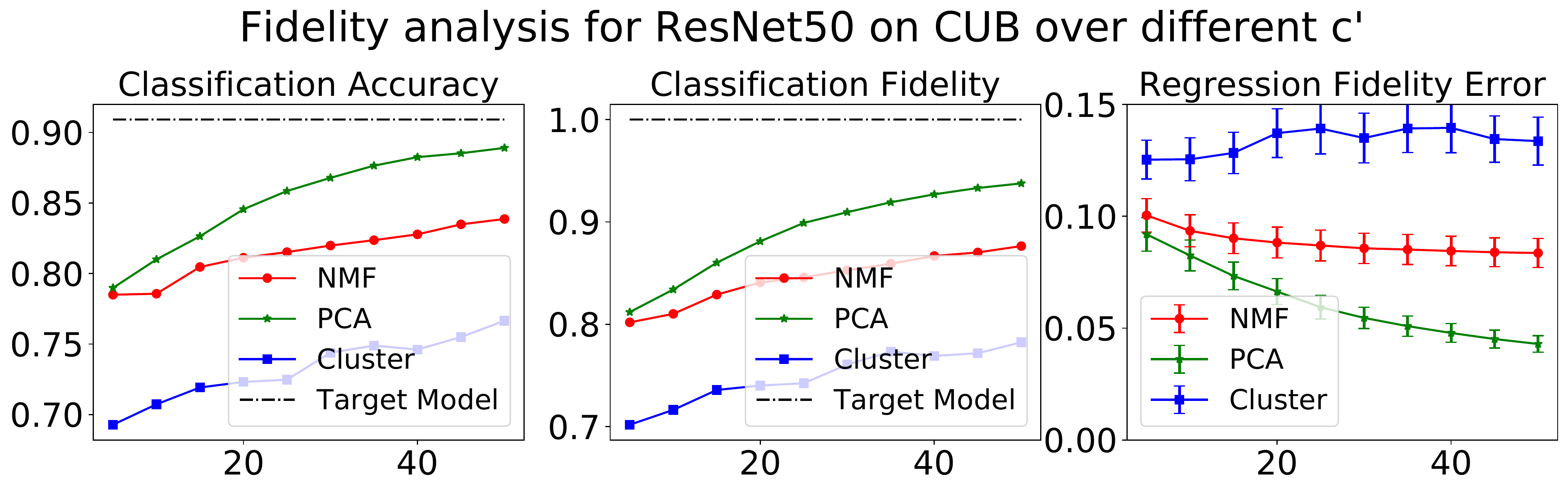}
    \caption{Average accuracy and fidelity for approximate linear models of two ResNet50 models over different number of concepts. Left figures show the accuracy if the approximate models are directly used as classifiers. For classification, higher means better. For regression, lower means better.}
    \label{fig:fidelity}
\end{figure}

\subsubsection{Experimental Results} 
Figure~\ref{fig:fidelity} shows the accuracy (left) and fidelity for different $c'$ with $Fid\_c$ (middle) and $Fid\_r$ (right). PCA provided the best fidelity result for both regression and classification. NMF's result is close to PCA's but diverges as $c'$ increases. PCA is a popular and efficient matrix factorization method. Compared to PCA, NMF has two limitations: non-negativity and no introduction of extra bias. Also, NMF finds new vector bases to achieve a new balance for each vector every time $c'$ increases, while PCA simply seeks a new basis vector iteratively, based on variance maximization. Clustering showed the worst performance. Clustering methods can be considered as matrix factorization methods, but they only provide one-hot vectors as centroid predictions offering the least information. When $c'$ increases, approximate models provide more faithful predictions for both classification and regression. Around 1,000 compute (8 core CPU, v100 GPU) hours are needed for the evaluation.

The fidelity results demonstrate the limitation of approximate models: explanations will be incorrect some of the time. However, the classification accuracy on the left shows a potential trade-off for domains where accurate explanations are important: we can throw away the final layers of the target model and use the approximate model instead. This will result in a 2-5\% drop in accuracy for NMF and PCA (30+ concepts), but the fidelity of the approximate model against itself is 100\%.


\subsection{Interpretability via Human Subject Experiments}

In this section, we present a new type of experiment for measuring the interpretability of concept-based explanations, and use this to evaluate the interpretability of approximate models based on NMF, clustering, and PCA. We agree with the claims by Leavitt and Morcos \cite{leavitt2020towards}
that interpretability should be measured in a falsifiable manner. In this paper, this includes measuring how well people can interpret and understand the concepts, rather than relying on ``intuition'' based on selected examples. These experiments provide a better basis for model comparison, compared to simply demonstrating interpretability via a handful of selected examples. To that end, we present a human-subject experiment design that measures the interpretability of concepts by asking study participants to provide free-text labels for concepts and measuring the consistency of labeling between participants.

We hypothesized that NCAVs learned from NMF are more interpretable than CAVs learned from clustering and PCA. We assume that CAVs learned from segments and feature maps using $k$-means are similar and consider CAVs from $k$-means clustering method on feature maps as ACE baselines, as both cluster feature maps and provide CAVs as outputs.

\subsubsection{Methodology:}
We use \emph{task prediction}~\cite[p.~11]{hoffman2018metrics} and the Explanation Satisfaction Scale~\cite[p.~39]{hoffman2018metrics} for evaluation. The assumption behind task prediction is that the ability to predict the output of a model implies a better understanding of how the model works.

\subsubsection{Experimental Design:} 
\begin{figure}[t]
    \centering
    \includegraphics[scale=0.28]{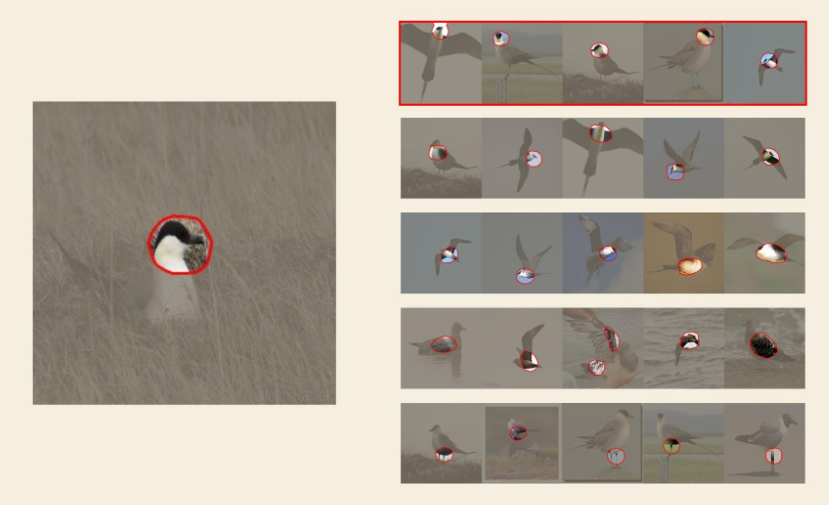}
    \caption{Survey trial sample in the prediction phase using NMF reducer. Participants instructions: ``\emph{Each row on the right reflects some concepts in the form of image samples generated by our AI model. Look carefully at the image on the left and select the concepts on the right which it most likely belongs to}.'' The first row is the ground truth.}
    \label{fig:survey}
\end{figure}

The experiment has two phases. In Phase 1, at each trial, participants are given one image with one concept highlighted and five concept explanations from the same class as candidates. Participants are asked to predict which of the candidate explanations matches the highlighted concept. An example is shown in Figure~\ref{fig:survey}. Then, for each concept candidate, participants are asked to provide a 1-2 word description of the concept. All participants are given five images as a training phase followed by 15 testing images. They can move back to a training example at any time in the test phase. In Phase 2, participants are asked to complete an explanation quality survey to self-report their opinion about explanations in the form of \citeauthor{hoffman2018metrics}'s explanation satisfaction scale. The experiment was implemented in a web-based environment on Amazon Mechanical Turk, a crowd-sourcing platform popular for obtaining data for human-subject experiments~\cite{buhrmester2011amazon}.

We used a between-subject design: participants were randomly assigned into one of nine groups (3 scenarios and 3 types of reducers). There were a total of 157 participants who completed the survey. Participants with a prediction accuracy lower than $20\%$ (random choice) were excluded from the results. Each experiment ran for approximately 30 minutes. We compensated each participant with \$5USD and an extra bonus of \$1USD for participants with high accuracy. $65\%$ of participants were male, $34\%$ were female and $1\%$ specified their own gender. Participants' reported ages were between 23 and 70 ($\mu = 38.9$).

Our measure for the task prediction is the percentage of correctly identified concept explanations. This assumes that better concept explainers allow participants to better match the highlighted concept to the correct explanation. For the 1-2 word descriptions,
participants should have similar descriptions for the concepts that correspond to clear meanings. We use GloVe~\cite{pennington2014glove} pre-trained word vector representations for each description, then use the average pairwise cosine similarity to measure the similarity of the concept descriptions across participants. We chose the free-text response task over providing pre-defined concept labels because using pre-defined labels could result in making concepts appear more interpretable than they are. If participants are uncertain, choosing the closest available label from a limited list makes the task about discrimination, rather than interpretation. Finally, we measure the participants' satisfaction with the explanations in terms of confidence, understanding, satisfaction, sufficiency and completeness \cite{hoffman2018metrics}. Explanations that are easy to interpret but perceived to be poor are unlikely to be used, so measuring subjective perception of explanations is important.
We obtained ethics approval from The University of Melbourne Human Research Ethics Committee (ID 1749428).

\subsubsection{Experimental Parameters:} 
To validate the consistency of results, we include three different scenarios: ResNet50 ($layer4$ as the target layer) for ILSVRC2012 (scenario RI), Inception-V3 ($Mixed\_7c$ as the target layer) for ILSVRC2012 (scenario II) and ResNet50 ($layer4$ as the target layer) for CUB (scenario RC) as target CNN models. The three methods NMF, Clustering and PCA are applied individually in each scenario.

The 20 images were drawn from 20 random classes chosen from all classes within a given dataset. For each class, we train an explainer with $c'$ of $10$, and only the top $5$ CAVs with the highest weights are chosen as selection candidates. One of these CAVs is randomly selected as the target. The concept in the target CAV is identifiable only if the sample image highly activates that CAV. So each target image is chosen from the top $10\%$ images in the test set which activates the target CAV most strongly (with high feature scores). Thus, we avoid using images in which the concept is absent (e.g., the \emph{tail} concept may be considered absent when only the upper part of a dog is shown in the image). Each CAV is visualized using the $5$ prototype samples. All explainers were trained with images from one class. The 20 classes were the same for different models for the same dataset, but candidate concepts and target instances were different.

\begin{table*}[t]
    \centering
    \caption{Top: Mean and standard deviation of prediction accuracy, description similarity, confidence and quality comparison for 9 different groups. Middle: ANOVA test p values for each scenario. Bottom: Bonferroni corrected T-test p values for each pair of reducers with significant ANOVA p value ($p < 0.05$) }
    
    \resizebox{0.94\linewidth}{!}{
    \begin{tabular}{ cccccccccc } 
        \toprule
        \multirow{3}{*}{\centering \textbf{Scenario}} & 
        \multirow{3}{*}{\centering \textbf{Reducer type}} &
        \multirow{3}{*}{\centering \textbf{Accuracy}} &
        \multirow{3}{50 pt}{\centering \textbf{Description Similarity}} &
        \multirow{3}{*}{\centering \textbf{Confidence}} &
        \multicolumn{4}{c}{\textbf{Quality}}\\
        \cmidrule(r){6-9}
        ~ & ~ & ~ & ~ & ~ & Understand & Satisfaction & Sufficiency & Completeness \\

        \midrule
        \multirow{3}{*}{RI} & NMF & \textbf{74.4\% $\pm$ 9.2\%} & \textbf{0.59 $\pm$ 0.1} & 77.7\% $\pm$ 13.0\% & \textbf{4.3 $\pm$ 0.6} & \textbf{4.1 $\pm$ 0.6} & 3.8 $\pm$ 0.8 & \textbf{3.7 $\pm$ 1.2} \\
        
        ~ & Cluster & 66.3\% $\pm$ 13.8\% & 0.56 $\pm$ 0.08 & 75.6\% $\pm$ 13.8\% & 4.2 $\pm$ 0.7 & 3.8 $\pm$ 1.0 & \textbf{3.9 $\pm$ 1.0} & 3.6 $\pm$ 1.1 \\
        
        ~ & PCA & 37.8\% $\pm$ 5.9\% & 0.52 $\pm$ 0.08 & \textbf{78.3\% $\pm$ 14.7\%} & 4.0 $\pm$ 0.9 & 3.8 $\pm$ 1.1 & 3.8 $\pm$ 1.1 & \textbf{3.7 $\pm$ 1.2} \\
        
        \midrule
        \multirow{3}{*}{II} & NMF & \textbf{62.6\% $\pm$ 18.6\%} & \textbf{0.57 $\pm$ 0.08} & 69.3\% $\pm$ 13.2\% & 3.5 $\pm$ 1.0 & 3.4 $\pm$ 1.3 & 3.3 $\pm$ 1.1 & 3.4 $\pm$ 1.3 \\
        
        ~ & Cluster & 44.8\% $\pm$ 13.2\% & 0.53 $\pm$ 0.09 & 75.1\% $\pm$ 13.7\% & \textbf{3.9 $\pm$ 1.1} & 3.6 $\pm$ 1.2 & \textbf{3.6 $\pm$ 1.2} & \textbf{3.5 $\pm$ 1.4} \\
        
        ~ & PCA & 40.0\% $\pm$ 8.6\% & 0.49 $\pm$ 0.08 & \textbf{76.0\% $\pm$ 13.0\%} & 3.8 $\pm$ 0.9 & \textbf{3.7 $\pm$ 1.1} & 3.4 $\pm$ 1.2 & 3.2 $\pm$ 1.3 \\
        
        \midrule
        \multirow{3}{*}{RC} & NMF & \textbf{81.1\% $\pm$ 8.4\%} & \textbf{0.7 $\pm$ 0.04} & \textbf{79.5\% $\pm$ 10.8\%} & \textbf{4.1 $\pm$ 0.8} & 3.7 $\pm$ 0.9 & 3.4 $\pm$ 1.2 & 3.5 $\pm$ 1.1 \\
        
        ~ & Cluster & 78.6\% $\pm$ 15.5\% & \textbf{0.7 $\pm$ 0.05} & 75.0\% $\pm$ 18.7\% & 3.9 $\pm$ 1.0 & \textbf{4.1 $\pm$ 1.0} & \textbf{4.0 $\pm$ 1.0} & \textbf{3.9 $\pm$ 1.1} \\
        
        ~ & PCA & 57.0\% $\pm$ 11.6\% & 0.59 $\pm$ 0.03 & 61.1\% $\pm$ 17.6\% & 3.6 $\pm$ 1.0 & 3.0 $\pm$ 1.2 & 3.4 $\pm$ 1.2 & 3.2 $\pm$ 1.2 \\
        \bottomrule

    \end{tabular}
    
    }

    \resizebox{0.94\linewidth}{!}{
   
    \begin{tabular}{ ccccccccc } 
        \toprule
        \multirow{3}{*}{\centering \textbf{Scenario}} & 
        \multirow{3}{*}{\centering \textbf{Accuracy}} &
        \multirow{3}{50 pt}{\centering \textbf{Description Similarity}} &
        \multirow{3}{*}{\centering \textbf{Confidence}} &
        \multicolumn{4}{c}{\textbf{Quality}}\\
        \cmidrule(r){5-8}
        ~ &  ~ & ~ & ~ & Understand & Satisfaction & Sufficiency & Completeness \\

        \midrule
        RI & \textless 0.001 & 0.131 & 0.841 & 0.446 & 0.592 & 0.941 & 0.948 \\
        
        II & \textless 0.001 & 0.064 & 0.304 & 0.493 & 0.752 & 0.690 & 0.844 \\
        
        RC & \textless 0.001 & \textless 0.001 & 0.004 & 0.283 & 0.016 & 0.219 & 0.174 \\
        \bottomrule
        
    \end{tabular}
    }    
    
    \resizebox{0.94\linewidth}{!}{
    \begin{tabular}{ cccccccccc } 
        \toprule
        \multirow{3}{*}{\centering \textbf{Scenario}} & 
        \multirow{3}{*}{\centering \textbf{Reducer Pair}} &
        \multirow{3}{*}{\centering \textbf{Accuracy}} &
        \multirow{3}{50 pt}{\centering \textbf{Description Similarity}} &
        \multirow{3}{*}{\centering \textbf{Confidence}} &
        \multicolumn{4}{c}{\textbf{Quality}}\\
        \cmidrule(r){6-9}
        ~ & ~ & ~ & ~ & ~ & Understand & Satisfaction & Sufficiency & Completeness \\

        
        
        
        
        
        
        
        

        \midrule
        \multirow{3}{*}{RI} & NMF vs. Cluster & 0.159 & \multirow{3}{*}{/} & \multirow{3}{*}{/} & \multirow{3}{*}{/} & \multirow{3}{*}{/} & \multirow{3}{*}{/} & \multirow{3}{*}{/} \\
        
        ~ & NMF vs. PCA & \textless 0.001 & ~ & ~ & ~ & ~ & ~ & ~ \\
        
        ~ & Cluster vs. PCA & \textless 0.001 & ~ & ~ & ~ & ~ & ~ & ~ \\
        
        \midrule
        \multirow{3}{*}{II} & NMF vs. Cluster & 0.019 & \multirow{3}{*}{/} & \multirow{3}{*}{/} & \multirow{3}{*}{/} & \multirow{3}{*}{/} & \multirow{3}{*}{/} & \multirow{3}{*}{/} \\
        
        ~ & NMF vs. PCA & \textless0.001 & ~ & ~ & ~ & ~ & ~ & ~ \\
        
        ~ & Cluster vs. PCA & 0.754 & ~ & ~ & ~ & ~ & ~ & ~\\
        
        \midrule
        \multirow{3}{*}{RC} & NMF vs. Cluster & 1.00 & 1.00 & 1.00 & \multirow{3}{*}{/} & 0.784 & \multirow{3}{*}{/} & \multirow{3}{*}{/} \\
        
        ~ & NMF vs. PCA & \textless 0.001 & \textless0.001 & 0.002 & ~ & 0.222 & ~ & ~ \\
        
        ~ & Cluster vs. PCA & \textless0.001 & \textless 0.001 & 0.088 & ~ & 0.023 & ~ & ~ \\
        \bottomrule
    \end{tabular}
    }

    \label{table:performance}    
\end{table*}

\subsubsection{Results}
Table~\ref{table:performance} shows the results of the human-subject experiment. We ran an ANOVA to evaluate the results and used pairwise t-tests to identify differences between each pair of reducers. In the prediction task, NCAVs from NMF are more interpretable than CAVs from PCA ($p < 0.001$), and CAVs from clustering are more interpretable than CAVs from PCA ($p < 0.001$). For description similarity, results are not significant ($p > 0.05$ level in most cases). Most CAVs contain some meaningful information; participants are confident about their choice. There is no significant difference in confidence and quality scores in most cases ($p > 0.05$). We conclude that NCAVs from NMF are more interpretable than CAVs from PCA. NCAVs are at least equally interpretable to CAVs from clustering.

\begin{figure}[t]
    \centering
    \includegraphics[scale=0.2]{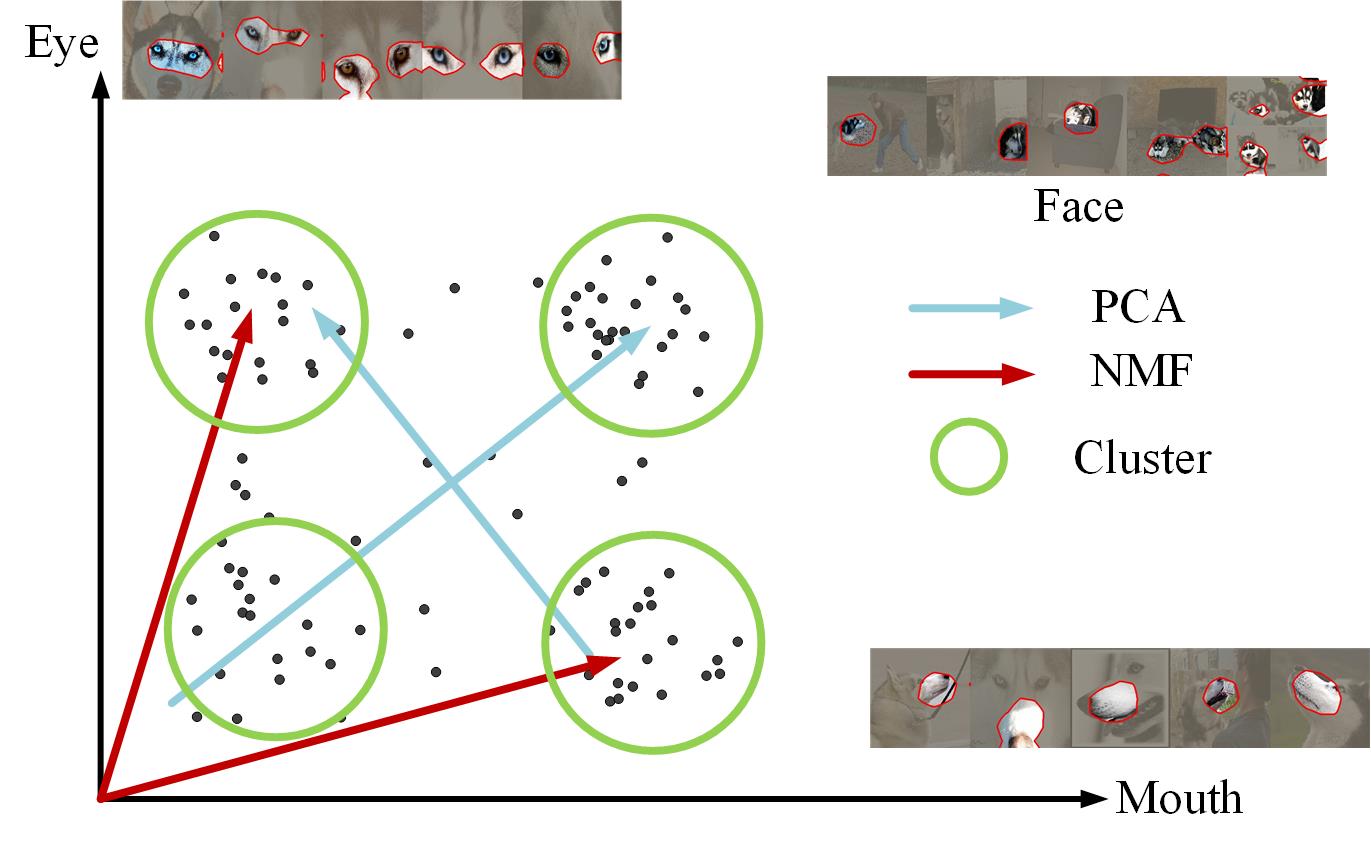}
    \caption{Having two concepts of `mouth' and `eyes' measured in two dimensions (only positive values reflect concepts), different reducers provide different directions to represent concepts. PCA learns less meaningful but efficient directions. Clustering methods could provide meaningful centroids' center directions but are the least efficient. NMF may provide meaningful directions with fewer dimensions.}
    \label{fig:assumption}
\end{figure}

We observe from our experiments that reducers can help generate meaningful concepts from feature maps, but fidelity and interpretability measure different aspects of performance. Here we propose an explanation for this phenomenon based on the differences between the three reducers. Figure~\ref{fig:assumption} shows a distribution of some concept instances. Each dot reflects an instance with some concept scores. Due to the $Relu$ activation in CNNs, we assumed that only positive values make sense in CNN models. The $X$ axis could contain the concept of `mouth' and $Y$ axis may reflect `eyes'. PCA has a new intersection of dimensions (bias) other than the root so one of the dimensions is meaningless (points to negative values). Clustering methods provide correct concept directions (from the root to the center of each cluster). But it may require more clusters for the same fidelity, since clustering is based on data clusters but not directions. Also, it may provide some similar concepts (bottom left and upper right clusters have similar directions). Clusters may also be influenced by some isolated instances and provide meaningless concepts. But for NMF, it provides correct concept directions in an efficient way if only positive values reflect meaningful concepts.

\section{Related Work}

This work focuses on explanations for pre-trained models. Common explanation methods provide explanations based on input level feature importance. Some methods provide model agnostic explanations based on importance for image segments~\cite{ribeiro2016should,lundberg2017unified} by training linear approximate models on nearby datasets. Saliency maps provide pixel-level feature importance for images based on gradients~\cite{shrikumar2017learning,bach2015pixel,smilkov2017smoothgrad}. However, some papers point out the unreliability of saliency methods~\cite{kindermans2017reliability,NIPS2019_9511}. CAM is another type of approach, providing heatmaps to indicate where the image activates the target class most based on CNN weights from the last several layers~\cite{zhou2016learning,selvaraju2017grad}. All these methods use features from input images.

Other than input level explanations, some papers build explanations from feature maps inside the CNN model and provide concept-level explanations based on supervised learning~\cite{kim2018interpretability,bau2017network,zhou2018interpretable}. These methods build linear classifiers on target concepts and use weights as CAVs. ACE~\cite{ghorbani2019towards} relaxes the limitation of the labeled dataset using unsupervised learning with $k$-means clustering. Learned concepts take the form of vectors from cluster centroids. 

Other than explanations for pre-trained models, some works modify the structure of CNN models to provide built-in interpretability~\cite{hendricks2016generating,zhang2018interpretable,chen2019looks}. Visualization of the values inside the CNN models can also help explain the process of prediction. Optimization methods can visualize which patterns produce the highest activation within the CNN models~\cite{olah2017feature}. There are many options for feature visualizations. These could choose different layers, different axes for target layer matrices, different matrices from CNN weights or instance feature maps and different methods of presentation to satisfy different interpretability requirements~\cite{olah2018the}. In this work, NMF is introduced to gather interpretable information from a feature map. NMF is a common interpretable matrix factorization method~\cite{lee1999learning},

\section{Conclusion}

We propose a framework for concept-based explanations for CNN models based on the post-training explanation method ACE. By using feature maps inside the CNN model, we can gather some interpretable concept vectors to provide explanations and invert them back to predictions. We also show that having requirements of fidelity and interpretability, NCAVs from NMF can provide overall better explanations compared with clustering and PCA methods. PCA provides CAVs with better fidelity but lack interpretability, which makes PCA less suitable for explanations. CAVs from clustering methods are interpretable but lower in fidelity.

\section*{Acknowledgements}
This research is supported by Australian Research Council (ARC) Discovery Grant DP190103414: \emph{Explanation in Artificial Intelligence: A Human-Centred Approach}. The first two authors are supported by the University of Melbourne research scholarship (MRS) scheme.
Experiments were undertaken using the LIEF HPC-GPGPU Facility hosted at the University of Melbourne. This Facility was established with the assistance of LIEF Grant LE170100200. 

\section{Ethics Statement}

This work focuses on improving the interpretability of CNN models, helping human analysts understand models' decision processes. Accountability and transparency are two key areas of responsible artificial intelligence \cite{dignum2019responsible}, and the ability to interpret models have a positive impact on both of these: accountability because making people accountable for decisions that are informed by machine learning is not reasonable if those people cannot gain insight as to why and how decisions are made; transparency because understanding the model helps to increase the transparency of decision making and to find inappropriate factors in the CNN models.

Despite this, there are potential negative impacts of this work. First, the explanations provided are approximate, and therefore can be incorrect in some cases. This has the potential to mislead people on the reasons for decisions. Second, examples are required as sample prototypes in the explanations. Inappropriate usage of training images may cause unintended and undesirable privacy disclosure if care is not taken to use public, consented sources or if privacy protections are not employed.

\bibliography{main}

\end{document}